\pdfoutput=1

\documentclass[11pt]{article}

\usepackage{ACL2023} 

\usepackage{times}
\usepackage{latexsym}

\usepackage[T1]{fontenc}

\usepackage[utf8]{inputenc}

\usepackage{microtype}
\usepackage{verbatim}

\usepackage{tabularx}
\usepackage{inconsolata}
\usepackage{todonotes}
\usepackage{enumitem}
\usepackage{hyperref}
\usepackage{nameref}
\usepackage{xcolor}
\usepackage{soul}
\usepackage{multirow}
\usepackage{nicematrix}
\usepackage{pifont}
\usepackage{arydshln}
\newcommand*\colourcheck[1]{%
  \expandafter\newcommand\csname #1check\endcsname{\textcolor{#1}{\ding{52}}}%
}
\newcommand*\colourcross[1]{%
  \expandafter\newcommand\csname #1cross\endcsname{\textcolor{#1}{\ding{55}}}%
}
\colourcheck{green}
\colourcross{red}

\renewcommand{\paragraph}[1]{\noindent{\textbf{#1}}}
\newcommand{\SideNote}[2]{\todo[color=#1,size=\small]{#2}} 
\newcommand{\violet}[1]{\SideNote{purple!40}{#1 --Violet}}

\title{Do Models Really Learn to Follow Instructions? An Empirical Study of Instruction Tuning}

\author{Po-Nien Kung, ~~~Nanyun Peng\\
   University of California, Los Angeles \\
   {\tt \{ponienkung,violetpeng\}@cs.ucla.edu } \\}

\begin{document}
\maketitle
\begin{abstract}

Recent works on instruction tuning (IT) have achieved great performance with zero-shot generalizability to unseen tasks. With additional context (e.g., task definition, examples) provided to models for fine-tuning, they achieved much higher performance than untuned models. 
Despite impressive performance gains, what models learn from IT remains understudied. 
In this work, we analyze how models utilize instructions during IT by comparing model training with altered vs. original instructions. 
Specifically, we create \textit{simplified task definitions} by removing all semantic components and only leaving the output space information, and \textit{delusive examples} that contain incorrect input-output mapping.
Our experiments show that models trained on \textit{simplified task definition} or \textit{delusive examples} can achieve comparable performance to the ones trained on the original instructions and examples. 
Furthermore, we introduce a random baseline to perform zero-shot classification tasks, and find it achieves similar performance ($42.6\%$ exact-match) as IT does ($43\%$ exact-match) in low resource setting, while both methods outperform naive T5 significantly (30\% per exact-match). 
Our analysis provides evidence that the impressive performance gain of current IT models can come from picking up superficial patterns, such as learning the output format and guessing. 
Our study highlights the urgent need for more reliable IT methods and evaluation.
\looseness=-1
\end{abstract}

\section{Introduction}

Recently, instruction tuning(IT) has drawn much attention in the NLP communities, with the rapid growth of new models~\cite{sanh2021multitask, wei2021finetuned, ouyang2022training} and datasets~\cite{Wang2022SuperNaturalInstructionsGV, gupta2022improving, finlayson2022makes, mishra2021cross, ye2021crossfit, bach2022promptsource}.
Models trained with task instructions demonstrate impressive zero-shot cross-task generalization ability. Despite the remarkable results, how models utilize the instructions during training and inference time remains an open question.

\begin{table}[t]
\centering
\small
\setlength{\tabcolsep}{1mm}
\begin{NiceTabular}{|l||ccc|cc|}
\Hline
\Block{2-1}{IT Models\\ models $\rightarrow$} & \Block{1-3}{Generalize to \\Unseen Tasks} &&& \Block{1-2}{Generalize to \\Unseen Instruct.} \\
    & TK-Inst & T0 & FLAN & Alpaca & Vicuna\\
\hline
Training & & & & &  \\
\# of tasks & 756 & 39 & 38 & \Block{1-2}{--} \\
\# of instructions & 756 &  390* & 380 & 52K & 70K\\
\hline
Testing & & & & &  \\
\# of tasks & 119 & 11 & 24 & \Block{1-2}{--} \\
\# of instructions & 119 &  110* & 240 & 252 & 252\\
\Block{1-1}{Testing on \\unseen tasks?} & \greencheck & \greencheck& \greencheck& \redcross & \redcross\\
\Hline
\end{NiceTabular}
\caption{\footnotesize Comparison between two types of instruction tuning models. Noted that we reported an estimated number of instructions for T0 during training and testing since they have 5 to 10 instructions for each task. Our analysis focuses on the ``generalize to unseen task'' type.}
\label{table:two-it-paradigms}
\vspace{-1.5em}
\end{table}

\begin{figure*}
    \centering
    \includegraphics[width=\textwidth]{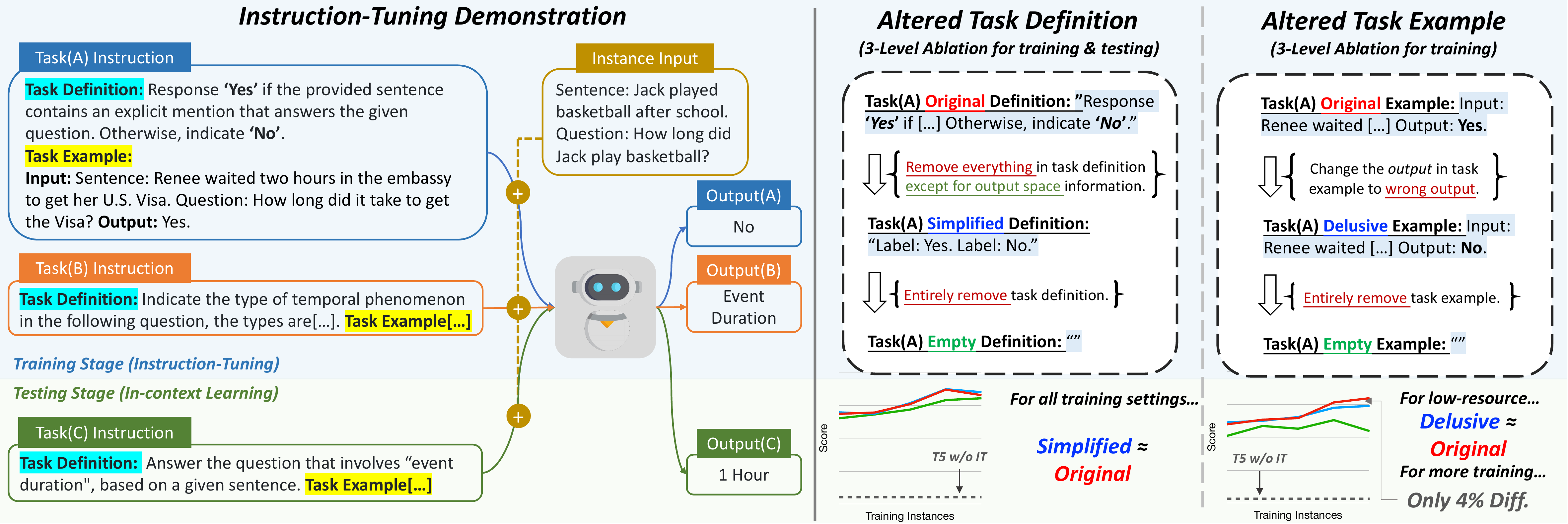}
    \vspace{-1em}
    \caption{\footnotesize The left sub-figure demonstrates a two-stage pipeline where the model first trains on a set of tasks and then evaluates other unseen tasks. The model inputs \textit{task definition}, \textit{examples}, and \textit{instance input} together to make a prediction.
    The two right sub-figures show how we create \textit{Simplified task definition} and \textit{Delusive task example} for ablation studies. We also demonstrate the results at the bottom with \textit{T5 w/o IT}(Untuned models) results. It is shown that models can still achieve significant performance gain compared to \textit{T5 w/o IT} while training on \textit{Simplified} task definition and \textit{Delusive examples}.
    }
    \label{fig:IT-Demo}
    \vspace{-1.5em}
\end{figure*}

Prior works have raised the question of whether models really learn to follow the instructions or just capture spurious correlations. 
\newcite{jang2022can}, \citet{webson2021prompt} showed that the current large language models (LLMs) can achieve similar performance with misleading instructions(prompts) in in-context learning(ICL) and few-shot learning scenarios.
\citet{min2022rethinking} analyze how model utilize examples in ICL. They observed that (1) Input-output mapping in examples is not important and(2) Output space information is crucial.

Besides ICL and few-shot prompt-tuning, some works raise concerns about instruction following in the instruction tuning field~\cite{finlayson2022makes, gupta2022improving, gu2022learning}, 
with a focus on test-time analysis. In contrast, we focus on analyzing how the models utilize instructions during the training process. We compare our analyzing methods and observation with prior works in Appendix \ref{appendix:analysis}. \looseness=-1
In this work, we conduct controlled experiments on NatInst-V2~\cite{Wang2022SuperNaturalInstructionsGV}, the largest open-source instruction learning dataset includes 800+ English tasks with diverse task types, to study how models utilize instructions during IT. 
Note that existing research on IT can be categorized into two major camps:
\textbf{generalize to unseen tasks} and \textbf{generalize to unseen instructions}, based on their objectives. \autoref{table:two-it-paradigms} shows the comparison. Our analysis focuses on the former with more background and justifications provided in \autoref{sec:background}.
We strategically alter the instructions and compare them with original instructions for IT. 
Specifically, for task definition, we create \textit{simplified versions} by removing all semantic components in the instructions and only leaving the output space information.
For task examples, we create \textit{delusive examples} with incorrect input-output mapping, where the examples’ input and output spaces are correct, but the input-output mappings are wrong. Figure \ref{fig:IT-Demo} demonstrates specific examples of these altered instructions.

Our experiments show that models trained with \textit{simplified task definitions} achieve performances on par with the original IT models with different numbers of training examples ranging from 10 to 800 per task.
We also observe that instruction-tuned models are sensitive to input-output mapping during the testing ICL stage, but not during the instruction-tuning (training) stage, especially in low resource settings (i.e., $\leq 50$ training instance per task).
To further understand why instruction tuning improves performance for zero-shot test tasks, we establish a random baseline that only knows the correct output format (label space) for classification and multi-choice tasks. 
We discover that 
the random baseline can get $30\%$ absolute exact-match score improvement over an untuned model, almost comparable to some IT models in low resource settings.\looseness=-1 

Our results suggest that 
the impressive performance gains of IT may just come from models learning superficial patterns, 
such as the output space and format. 
We suggest future research on IT more carefully analyze their performance gains and benchmark against trivial baselines. 

\section{Background}
\label{sec:background}
Recently, many instruction tuning work train and test the models with instructions to achieve better zero-shot generalizability toward unseen tasks/instructions. We categorize these works by their objectives:
\textbf{generalize to unseen tasks} and \textbf{generalize to unseen instructions}, and show the comparison in \autoref{table:two-it-paradigms}. 

\paragraph{Instruction tuning to generalize to unseen tasks.}
Figure \ref{fig:IT-Demo} illustrates a two-stage instruction tuning pipeline used in many IT models, such as T0~\cite{sanh2021multitask}, FLAN~\cite{wei2021finetuned}, and TK-Instruct~\cite{Wang2022SuperNaturalInstructionsGV}.
In the first stage, the models are trained on a set of training tasks with instructions (task-definition and task-examples). After training, the models are evaluated on a set of unseen testing tasks for zero-shot generalizability.
By incorporating instructions during training, the models are shown to significantly improve performance over untuned models. The impressive performance gains led people to believe that models learned to follow instructions via instruction tuning. 
The goal of our analysis is to verify this belief. 

\paragraph{Instruction tuning to generalize to unseen instructions.}
Different from T0, FLAN, and TK-Instruct training and testing the model with clear task boundaries and focusing on cross-task generalizability, Instruct-GPT~\cite{ouyang2022training}, Alpaca~\cite{alpaca} and Vicuna~\cite{vicuna2023} focus more on instruction generalizability, which they train their model without clear task boundary but with diverse instructions, and further test on user-oriented instructions.
These models show very different behavior compared with instruction tuning models that aim to generalize to unseen tasks. \looseness=-1

Since Instruct-GPT is not open-sourced and distilled IT models such as Alpaca and Vicuna come up after our submission, we focus our analysis on the first category using the TK-instruct model and NatInst-V2 dataset. 
However, we also conduct additional experiments and discuss the Alpaca model's instruction following ability in \autoref{table:alpaca}.




\section{Analysis Method}
\paragraph{Task definition manipulation.}
To analyze whether models really ``understand'' and utilize the semantic meaning of task definitions, we conduct controlled experiments to remove semantic information in task definitions.
Specifically, we conduct instruction-tuning with task definitions at 3 levels of granularity: \textbf{Original}, \textbf{Simplified}, and \textbf{Empty}. 
The \textbf{Original} version uses human-crafted human-readable task definitions provided in NatInst-V2 \cite{Wang2022SuperNaturalInstructionsGV}.
The \textbf{Simplified} task definitions remove all semantic components in the original task definition and only leave the output space information. Specifically, we only provide possible output labels as task definitions for classification tasks, 
and completely remove task definitions for other tasks (mostly generative tasks) during IT. 
Figure~\ref{fig:IT-Demo} shows an example of \textbf{Simplified} task definition. More details can be found in Appendix \ref{appendix:ablate-examples}.
For \textbf{Empty}, we don't provide task definition during instruction-tuning. \looseness=-1


\paragraph{Task example manipulation.}
\citet{finlayson2022makes} show that by 
providing a few task examples, 
both humans and models can guess and perform a task.
We thus design a controlled experiment to study whether models learn the input-output mapping from task examples.
Specifically, we compare models trained with 3 types of task examples: \textbf{Original}, \textbf{Delusive}, and \textbf{Empty}.
For the \textbf{Original} setup, we provide one positive example in NatInst-V2~\cite{Wang2022SuperNaturalInstructionsGV}.
For \textbf{Delusive} examples, we sample negative examples from NatInst-V2, which have correct input and output formats, but incorrect input-output mappings.
For \textbf{Empty}, we do not provide task examples during training.
 

\begin{figure}[t!]
    \centering
    \includegraphics[width=0.5\textwidth]{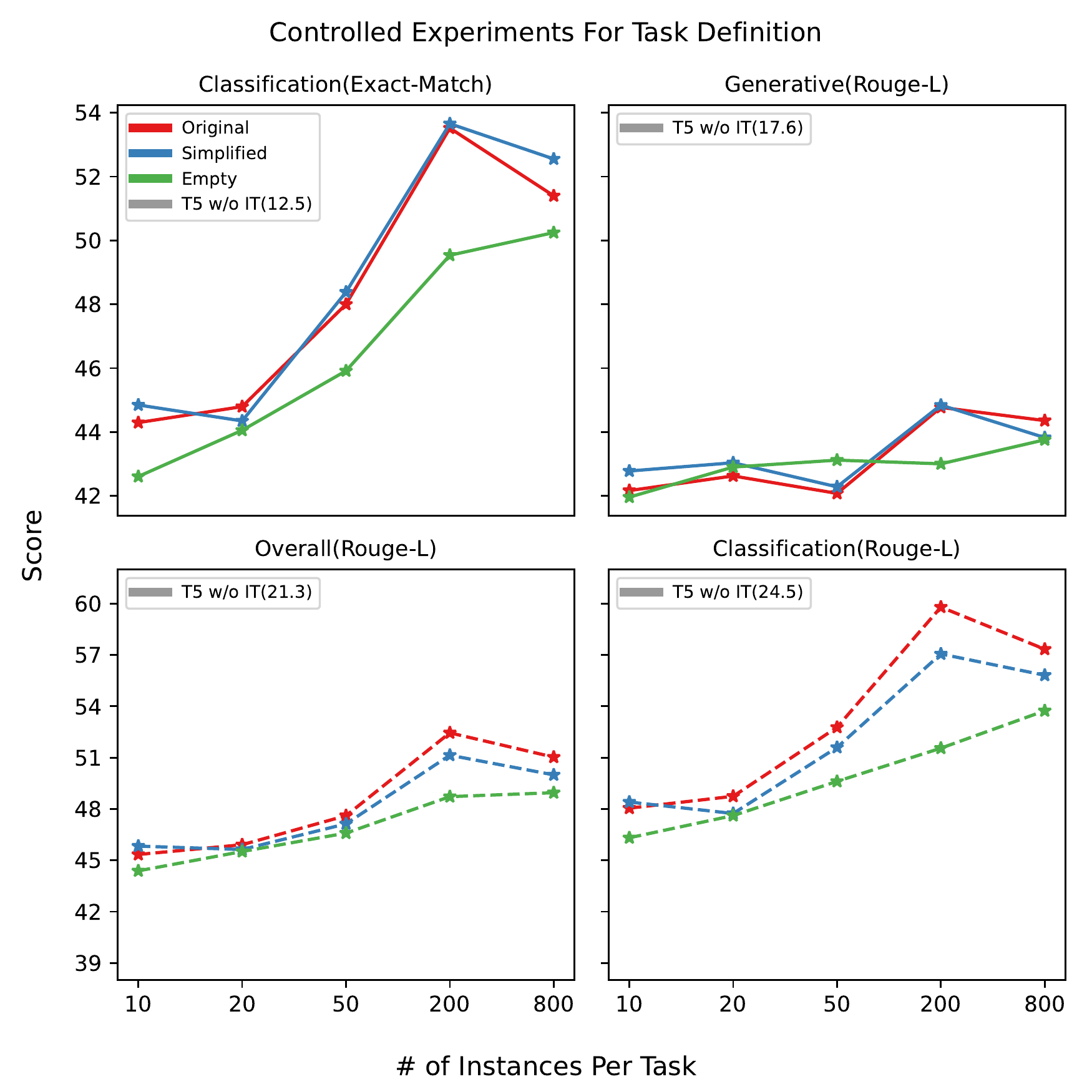}
    \vspace{-2em}
    \caption{\footnotesize Controlled experiments for task definition. \textbf{Original}, \textbf{Simplified}, and \textbf{Empty} represent what type of task-definition the model is trained and tested with. \textbf{T5 w/o IT(12.5) shows the score(12.5) of the baseline T5-large model.}
    The top two subfigures show the main results evaluating classification tasks using Exact-Match (accuracy) and Generative tasks using Rouge-L. The bottom two sub-figures are supplementary results evaluating rouge-L for \textit{All tasks} and \textit{classification tasks}.}
    \label{fig:task-def-results}
    \vspace{-1.5em}
\end{figure}

\section{Experimental Setup} 
\paragraph{Dataset.}
We conduct experiments on the NatInst-V2 \cite{Wang2022SuperNaturalInstructionsGV}, the largest open-source instruction learning dataset, including over 800+ English tasks with diverse task types. The instructions include human-crafted human-readable \textit{Task Definition}, \textit{Positive Task Examples}, \textit{Negative Task Examples}, and \textit{Explanation}. We focus on studying task definition and task examples, which were shown to be most useful in the original paper. 

\paragraph{Model.} we conduct experiments on TK-Instruct, the current SOTA model provided in NatInst-V2 paper. The model significantly outperformed previous SOTA models, such as T0 (62.0 v.s. 32.3 rouge-L for 11B model). 
We follow the seq-to-seq instruction-tuning method used in TK-Instruct, and train a T5-large-lm-adapt (770M parameters) model \cite{raffel2020exploring} with performance comparable to the larger model (3B parameters) reported in \newcite{Wang2022SuperNaturalInstructionsGV}.\footnote{For task definition experiment, we follow the best performance settings from \citet{Wang2022SuperNaturalInstructionsGV} to use task definition and two examples as instructions. For task examples experiments, due to the lack of negative examples, we conduct ablation studies using task definition and one example.}\looseness=-1
\begin{figure}
    \centering
    \includegraphics[width=0.5\textwidth]{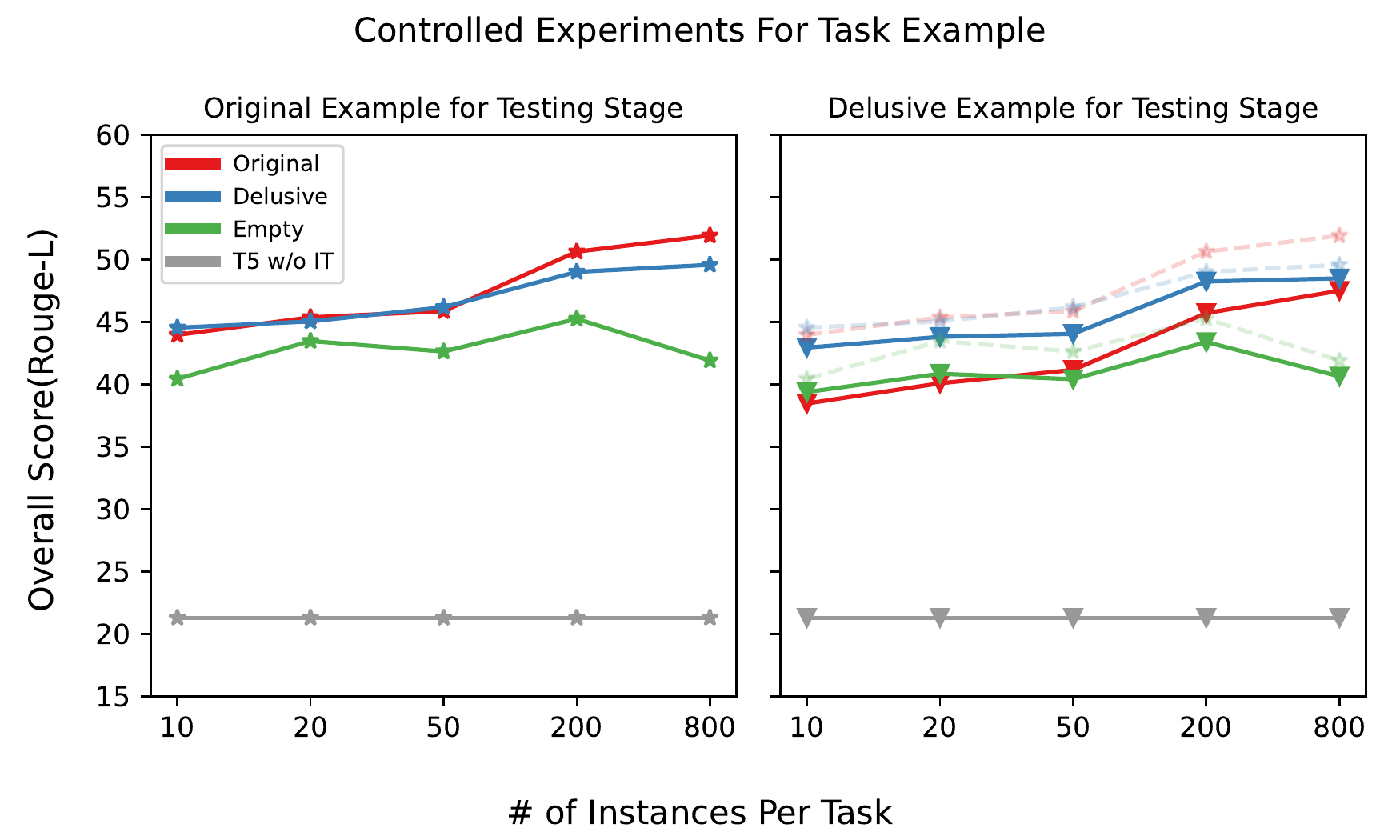}
    \vspace{-2em}
    \caption{\footnotesize Controlled experiments for task examples. The left sub-figure shows the main results, where \textbf{Original} task examples are used for testing (in-context learning). \textbf{Original}, \textbf{Delusive}, and \textbf{Empty} represent what type of task examples are used for training and the \textbf{T5 w/o IT} is the baseline T5-large model.
    The right sub-figure shows supplementary results using \textbf{Delusive} examples for testing. The faint dashed lines are copied from the left sub-figure for comparison purposes.
    }
    \label{fig:task-example-results}
    \vspace{-1.5em}
\end{figure}

\paragraph{Evaluation Metrics.}
For task definition, we separately evaluate \textit{Classification} and \textit{Generative} tasks using exact match and rouge-L respectively. 
For task examples, we follow~\newcite{Wang2022SuperNaturalInstructionsGV} to report the overall rouge-L score for both classification and generative tasks. 
To understand the impact of training examples, we report model performances with varying numbers of training instances per task (i.e., 10, 20, 50, 200, 800). 

\section{Results} 
\label{results}
\paragraph{Task Definition Experiments.}
Figure \ref{fig:task-def-results} shows experimental results for task definitions. 
In the top sub-figures, we can see that the models trained with \textbf{Simplified} instructions achieve almost the same results as models trained with \textbf{Original} definitions both on Classification and Generative tasks.
Note that \textbf{Simplified} task definitions remove all semantic components in task definitions and only retain output space information for Classification tasks and remove task definitions altogether for Generative tasks. 
This indicates that models may only utilize output space information during instruction tuning.
The bottom-left sub-figure in Figure \ref{fig:task-def-results} shows the overall rouge-L score for classification tasks, where models trained on the \textbf{Original} task definition slightly outperform the \textbf{Simplified} ones.  
A closer examination reveals that models trained on the \textbf{Original} task definitions are more likely to predict partially correct answers that help with the ROUGE-L score in some tasks. We provide further details in Appendix \ref{appendix:perf-gap}.
In addition, we also observe that training with \textbf{Simplified} prompts can yield comparable performance to the T0 model trained with \textbf{Original} prompts on T0 dataset. Please refer to Appendix \ref{appendix:t0} for details.

\begin{figure}
    \centering
    \includegraphics[width=0.5\textwidth]{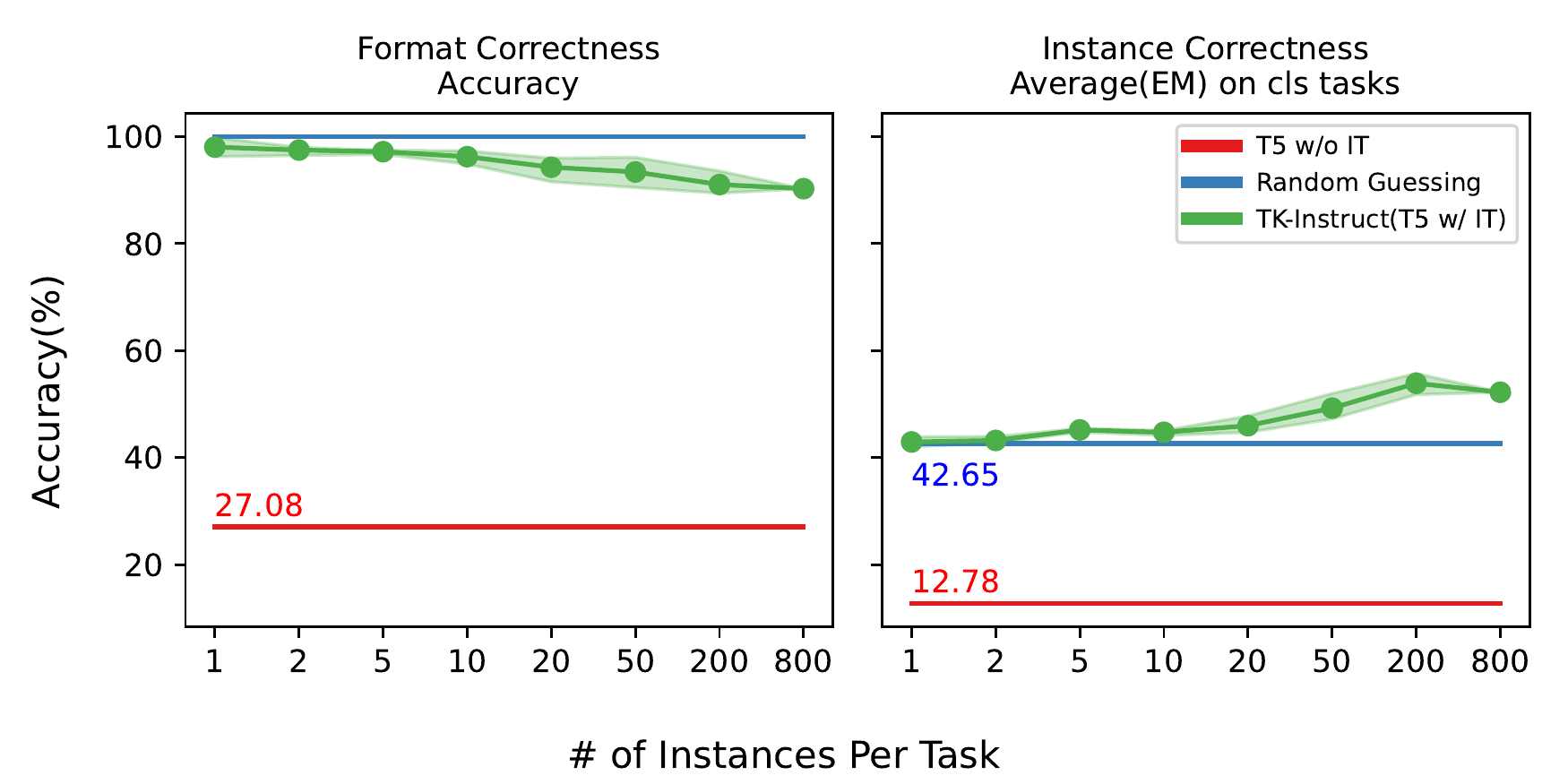}
    \vspace{-2em}
    \caption{\footnotesize Results for the \textbf{Random Guessing} baseline which randomly guesses an answer from the output space (labels). The left figure shows the format correctness, which calculates the accuracy of model predictions lied in the label space for classification (CLS) tasks. 
    The right figure shows the average exact-match score of CLS tasks. 
    }
    \vspace{-1.5em}
    \label{fig:format-correct}
\end{figure}

\paragraph{Task Examples Experiments.}
Figure \ref{fig:task-example-results} shows the experimental results for task examples. 
The left sub-figure shows overall ROUGE-L scores. It shows that models trained with \textbf{Delusive} task examples can achieve almost the same performance as \textbf{Original} task examples when the number of training instances per task is small ($\leq 50$). When the data per task goes to 200, the \textbf{Original} models started to outperform \textbf{Delusive} ones slightly. 
Combined with the previous results for task definition, we observe that comparing to the untuned models(\textit{T5 w/o IT}), the IT models may achieve significant performance gain (Rouge-L from 22 to 46) with (1)\textit{Simplified} task definition and (2)\textit{Delusive} task example, indicating that the current impressive improvement of IT models can come from the models learning superficial patterns without utilizing (following) the instructions like human do.

For the right sub-figure, we show the results using \textbf{Delusive} task examples during test time via in-context learning. We see the performance drops for all three models, indicating that the input-output mapping matters for in-context learning on instruction-tuned models. 
This observation seems to misalign with previous work~\cite{min2022rethinking}, which they found input-output mapping is unimportant for in context learning for classification tasks. 
However, a closer investigation found that most tasks suffer from significant performance drop are analogical tasks rather than classification tasks as studied in \newcite{min2022rethinking}.\footnote{See examples of analogical tasks in Appendix \ref{appendix:analogical}.} 

\section{Additional Analysis}
\label{sec:analysis}
\paragraph{Random baseline.}
While our experiments suggest that models do not utilize most information in the instructions, we still observe huge performance gains via instruction tuning. 
To understand where the gains come from, we introduce a \textbf{Random} baseline that simply guesses within the correct output space.  
Figure \ref{fig:format-correct} shows the results.
First, IT improves format correctness from $27\%$ to $97\%$ by training with only one instance per task, and the exact-match score improves from $12.78\%$ to $43\%$. Further providing more training instances per task($200$) can improve exact-match score to $52\%$.
However, while the performance gains seem impressive, the \textbf{Random Guessing} baseline can also achieve $42.6\%$ exact-match score, on par with TK-Instruct trained in low resource setting (less than five instances per task).
This suggests that the majority of score improvement from IT may come from model learning the output format, especially in low-resource settings.

\begin{table}[t]
\centering
\small
\setlength{\tabcolsep}{1.5mm}
\begin{NiceTabular}{|l|cccc|}
\Hline
Model / Metric & \Block{1-1}{CLS\\(EM)} & $\Delta $& \Block{1-1}{GEN\\(Rouge-L)} & $\Delta$\\
\Hline
LLaMA & &\\
\textit{Test w/} \textbf{Original}  & 4.40 & & 14.31 &\\
\hdashline
\textit{Train w/} \textbf{Original} & 59.19 & \Block{2-1}{-2.58}&48.80 & \Block{2-1}{-3.05}\\
\textit{Train w/} \textbf{Simplified} & 56.61 & &45.75 &\\
\Hline
Alpaca &&\\
\textit{Test w/} \textbf{Original} & 45.08 & \Block{2-1}{-3.42} & 44.40 & \Block{2-1}{\textbf{-9.6}}\\
\textit{Test w/} \textbf{Simplified} & 41.66 & &34.80 &\\
\hdashline
\textit{Train w/} \textbf{Original} & 59.33 & \Block{2-1}{-3.16} &48.69&\Block{2-1}{-3}\\
\textit{Train w/} \textbf{Simplified} & 56.17& & 45.69&\\
\Hline
\end{NiceTabular}
\caption{\footnotesize Altered task definition experiment for LLaMA and Alpaca model. \textbf{Original} and \textbf{Simplified} specify the provided task definition type. \textit{Test w/} specify how the model is tested with provided task definition type. \textit{Train w/} specify how the model is trained and tested with provided task definition type. We provide task definition and one example as instruction for both training and testing.}
\label{table:alpaca}
\vspace{-1.5em}
\end{table}

\paragraph{Fair comparison for IT models.} 
Existing studies on instruction tuning often introduce changes to both models and datasets simultaneously, which can obscure fair comparisons. To address this issue, we conduct experiments comparing different models (T0, TK-Instruct) on the same dataset (NatInst-V2) and emphasize the importance of careful evaluation.
In Table \ref{table:careful-eval}, when evaluating using the NatInst-V2 evaluation method and considering only the overall Rouge-L score, the TK-Instruct model appears to outperform T0 significantly. However, upon closer examination of the classification (CLS) and generative (GEN) tasks separately, we observe that T0's classification score is even lower than the Random baseline, primarily due to its format correctness being only 64\%.
To ensure a fairer comparison between these models, we employ constrained decoding techniques to align the model's predictions with the label space. By adopting this approach, we observe a substantial performance improvement for T0 in CLS tasks (34.03 to 51.31). T0 surpasses both the TK-Instruct model and the random baseline, indicating that it is indeed superior to these models in CLS tasks.
\looseness=-1
\begin{table}[t]
\centering
\small
\setlength{\tabcolsep}{1.5mm}
\begin{NiceTabular}{|l||c|ccc|}
\Hline
\Block{1-1}{Metric $\rightarrow$\\Model $\downarrow$} & \Block{1-1}{Format\\(Acc)} & \Block{1-1}{CLS\\(EM)} & \Block{1-1}{GEN\\(Rouge-L)} & \Block{1-1}{Overall\\(Rouge-L)}\\
\Hline
Random & 100 & 42.65 & -- & -- \\
\Hline
T0 & 64.61 & 34.03 &27.36 &32.28 \\
w/ CD & 100 & \textbf{51.31} & 27.36 & 40.7 \\
\Hline
TK & 96.23 &44.29 &\textbf{42.16} &45.34 \\
w/ CD & 100 & 47.12 & \textbf{42.16} & \textbf{45.93}\\
\Hline
\end{NiceTabular}
\caption{\footnotesize Careful evaluation of the NatInst-V2 dataset. The \textit{Format} metric is the same as the format correctness in \autoref{fig:format-correct}. The \textbf{w/ CD} indicates that the model's decoding is constrained to match the label choices for CLS tasks. The \textbf{TK} is the TK-Instruct(770M) model trained with 10 instances per task.}
\label{table:careful-eval}
\vspace{-1.5em}
\end{table}

\section{Discussion}
\label{sec:discussion}

\paragraph{Do Alpaca better follow the instruction on NatInst-V2 dataset?}
\label{para:alpaca}
After our submission, new instruction tuning models, like Alpaca and Vicuna, are trained on distilled data from Chat-GPT and exhibit behavior closer to it.
To investigate their instruction utilization, we conduct the ``Altered Task Definition'' experiment on LLaMA-7B~\cite{touvron2023llama} and Alpaca-7B models using the NatInst-V2 test set.
In  Table \ref{table:alpaca}, training the LLaMA model on the NatInst-V2 dataset using the \textbf{Original} task definition leads to substantial performance enhancements than zero-shot. However, the \textbf{Simplified} task definition also achieves comparable performance, with a minimal decrease of 3 (EM/Rouge-L)scores. This finding is consistent with our previous observations on the TK-Instruct and T0 models.
Even without tuning on NatInst-V2, the Alpaca model demonstrates strong performance on the NatInst-V2 test set. However, when the model is tested using a \textbf{simplified} task definition, there is a significant decrease in performance for generative tasks (but not for classification tasks). This highlights the importance of a well-written task definition for the Alpaca model to effectively perform generative tasks.\looseness=-1

\section{Conclusion}
We constructed controlled experiments on NatInst-V2 to compare model training with altered vs. original instructions (task definitions and examples). 
Our findings indicate that some current IT models do not fully utilize instructions, and the impressive performance gains of IT may come from models learning superficial patterns, such as the output space and format.
We suggest future research on instruction tuning to analyze their performance gains with more comprehensive evaluation and benchmark against trivial baselines.


\clearpage

\section{Limitations}
While our analysis suggests that IT models do not fully utilize instructions but instead learn superficial patterns from instructions, there are some limitations to our experiments.
First, we only analyze a SOTA IT method on the NatInst-V2 dataset and T0 dataset. Though \citet{Wang2022SuperNaturalInstructionsGV} showed that their model can outperform other large models such as Instruct-GPT~\cite{ouyang2022training} and T0~\cite{sanh2021multitask}, we did not analyze other IT methods, such as RLHF (Reinforcement Learning from Human Feedback) in Instruct-GPT.
Secondly, since our analysis is conducted in the training stage, we cannot analyze private models such as Chat-GPT. Also, we did not explore models larger than 7B parameters due to our computation resource limitation. This may miss some emergent abilities of large language models (LLMs)~\cite{wei2022emergent}. 
Lastly, while we observe the models do not utilize the majority of the instructions by IT, a certain degree of instruction understanding may already exist in pre-trained LLMs, which we did not study in this work.
In conclusion, our work is a concentrated analysis to illuminate the potential vulnerability of the current IT models and evaluation metrics. We encourage future works to conduct more comprehensive studies on larger models and propose more reliable IT methods and evaluation frameworks.

\section{Ethical Considerations}
We will go through the computation resources and models we used to conduct our experiments.
All of our models run on 4 48GB NVIDIA A6000 GPUs, along with 48 TB disk storage and AMD EPYC 7413 24-Core Processor. The experiment take around 1200 GPU hours for one 48GB NVIDIA A6000 GPU. Our experiments do not need to leverage model or data parallelism.
For the model, we use Huggingface T5-large-lm-adapt models for
our experiments, and will release our code once the paper been accepted.

\section*{Acknowledgements}
Many thanks to Zefan Cai for implementing the altered task definition experiment on the T0 dataset and model.
We would also like to thank Te-Lin Wu and Da Yin for their valuable insights during discussion, paper reviews, and constructive comments.
We thank the anonymous reviewers for their feedback.
This work was partially supported by AFOSR MURI via Grant \#FA9550- 22-1-0380, Defense Advanced Research Project Agency (DARPA) grant \#HR00112290103/HR0011260656.
\bibliography{anthology,custom}
\bibliographystyle{acl_natbib}

\appendix

\section{Appendix}
\subsection{Related Analysis.}
\label{appendix:analysis}
\citet{min2022rethinking} found input-output mapping in examples is irrelevant for in-context learning (ICL) on \textit{classification tasks}. However, we observe that it matters to ICL but is irrelevant to IT training on 
\textit{analogical generative tasks}.  
\citet{webson2021prompt} analyzed prompt-based models in few-shot learning scenarios and observed that models learn as fast using irrelevant or misleading prompts, which aligned with our findings. 
For instruction tuning, prior works raised concerns about models not following instructions. \citet{gu2022learning, gupta2022improving} analyze how models utilize instructions by removing them during inference stages. However, they did not address how models use instructions during training. \citet{wei2021finetuned, Wang2022SuperNaturalInstructionsGV} observe performance drop when removing task definition during IT and conclude that task definition is helpful, which we found true but only in terms of providing output space information.
Additionally, a concurrent study~\cite{yin-2023-did} has undertaken a comprehensive analysis of how models employ task definition in the process of instruction tuning on the NatInst-V2 dataset. They observed that by removing a majority of components from the task definition and retaining only the user intent, the model can attain comparable or even superior performance compared to utilizing the complete task definition.
\subsection{Simplified Task Definition}
\label{appendix:ablate-examples}
To remove all semantic components and only leave the output space information within the task definition, we first manually look through all tasks to verify how each task definition describes their output space and further categorize all task definitions into four types: \textbf{(1) Exact Mentioned, (2) Combined Mentioned, (3) Keyword Mentioned, and (4) No Mentioned}.
For \textbf{Exact Mentioned, Combined Mentioned} and \textbf{Keyword Mentioned}, there is a description of output space in the original task definition. For \textbf{No Mentioned}, The original task definition doesn't directly describe the labels or keywords in output space. This includes all the generative tasks and some classification tasks(We observe a few classification tasks in which task definitions do not describe output space information). Further details and examples are shown in Table \ref{table:Example-Task-Def}. \\

\subsection{Hyper-parameter tuning results}
\label{appendix:hyper}
Before we conduct analysis, we follow the model settings in \citet{Wang2022SuperNaturalInstructionsGV} to perform the hyper-parameter search. Prior works trained the TK-Instruct(770M) models from T5-Large-lm-adapt(770M) with a learning rate 1e-5, batch size 16, and 100 training instances per task for two epochs. We found out that (1) learning rate 1e-4 can converge faster while performance remains; (2) Higher batch size($\geq 128$) leads to much lower loss and better performance; (3) more training instances per task($\geq 200$) leads to better performance; and (4) the loss will converge with 4 to 6 epochs.
Following the hyper-parameter search results, we conducted our experiment with the following setting: learning rate 1e-4, batch size 128, [10, 20, 50, 200*, 800] training instance per task, and trained for six epochs. Our best results(200 instances) achieve a 52.8 Rouge-L score, which is better than TK-Instruct-770M(48 Rouge-L) from \citet{Wang2022SuperNaturalInstructionsGV} and comparable to their TK-Instruct-3B(54 Rouge-L) model.
\subsection{Analogical Tasks}
\label{appendix:analogical}
We look into a set of models training with \textit{Original} task examples and find out a list of tasks with the most performance drop(Drop more than 20\% score) when using \textit{Delusive} examples during testing(in-context learning). We show the list of tasks in Table \ref{table:analogical-task-list} and some of their details in Table \ref{table:analogical-example}. It is seen that these types of tasks have short input and output lengths, where input and output have direct word-level relations.

\subsection{Performance gap between rouge-L and exact match}
\label{appendix:perf-gap}
In the Results section, we observed that there's a slight performance gap on \textit{Classification} tasks between model training with \textit{Original} and \textit{Simplified} task definition. By further examining the data, we observed that this could happen to some \textit{Keyword Mentioned} tasks we described in Appendix \ref{appendix:ablate-examples}. Table \ref{table:Example-Task-Def} shows the example tasks in \textbf{Keyword Mentioned}. This task is a 7-class classification task with a special label "REFERENCE". The ground truth with "REFERENCE" will be combined with other text in the input, and both \textit{Original} and \textit{Simplified} models struggles(0\% exact match) to predict the correct answer for this class. However, while both models failed to predict exactly correct answers, we observed that the \textit{Original} model could achieve better partially correct answers by simply predicting more "REFERENCE". When we look into the testing set, we observe that 94 percent of ground truth is in "REFERENCE" class.
Also, when we look into the predictions, we observe \textit{Original} model will predict 55 percent of "REFERENCE" while \textit{Simplified} only predicts 4 percent, achieving a 33.8 higher rouge-L score. We hypothesized that this happened because the word "reference" has explicitly been mentioned numerous times(8) in the \textit{Original} task definition while mentioning other labels less than twice, leading to \textit{Original} model's tendency to predict "REFERENCE".\looseness=-1

\subsection{Simplified task definition for T0.}
Besides analyzing on NatInst-V2 dataset, we also conduct the simplified task definition experiment on T0 training stages. We follow the T0 training settings and changed the prompts to \textbf{Simplified} prompt, leaving only labels in the prompt for classification tasks and removing the entire prompt for generative tasks. 
We further train the T0-3B model using learning rate 1e-4, batch size 1024 for 10000 steps.
The T0 model training and testing with \textbf{Simplified} prompts achieve a 60.69 overall score, which is comparable to training with \textbf{Original Prompt}(61.93) and aligns with our observation on the NatInst-V2 dataset.
\label{appendix:t0}



\begin{table*}
\centering
\small
\setlength{\tabcolsep}{2mm}
\begin{tabularx}{\linewidth}{|l|X|} 
 \hline
 \multicolumn{2}{|c|}{Exact Mentioned}\\
 \hline
 Description & 
For tasks labeled as \textbf{Exact Mentioned}, the task definition describes the finite output space, which means all the labels within the output space are directly written in the definition.
 \\
 \hline
 Original Definition &
 Definition: In this task, you will be shown a short story with a beginning, two potential middles, and an ending. Your job is to choose the middle statement that makes the story incoherent / implausible by indicating \hl{1} or\hl{2} in the output. If both sentences are plausible, pick the one that makes less sense.
 \\ \hline
 Output Space &
 Finite Set: ["1", "2"]
 \\ \hline
 Simplified Definition &
 "Label: 1. Label: 2." \\ \hline\hline
 \multicolumn{2}{|c|}{Combined Mentioned} \\
 \hline
 Description &
 For tasks labeled as \textbf{Combined Mentioned}, the task definition describes a set of keyword labels that construct an infinite output space with all possible combinations of these keyword labels.
 \\
 \hline
 Original Definition &
 Given a command in a limited form of natural language, provide the correct sequence of actions that executes the command to thus navigate an agent in its environment. [...] 
 There are only six actions: 
 , \hl{'I\_WALK'}, \hl{'I\_RUN'}, \hl{'I\_JUMP'}, \hl{'I\_TURN\_LEFT'}, and \hl{'I\_TURN\_RIGHT'}. [...]
 \\ \hline
 Output Space &
 Infinite Set: ["I\_LOOK", "I\_LOOK I\_WALK", "I\_JUMP I\_RUN", ... $\infty$]
 \\ \hline
 Simplified Definition&
 "Combined Label: I\_LOOK. Combined Label: I\_WALK. Combined Label: I\_RUN. Combined Label: I\_JUMP. Combined Label: I\_TURN\_LEFT. Combined Label: I\_TURN\_RIGHT."
 \\ \hline\hline
  \multicolumn{2}{|c|}{Keyword Mentioned} \\
 \hline
 Description &
 For tasks labeled as \textbf{Keyword Mentioned}, 
 the task definition describes a set of keyword labels that construct an infinite output space combined with the input text.
 \\
 \hline
 Original Definition &
 In this task, you will use your knowledge about language (and common sense) to determine what element the marked number refers to. [...] Options to choose from are: \hl{REFERENCE}: Some object which is being mentioned in the text before or after the target number. The reference answer has a higher priority than any other. If both Reference and another answer are possible, prioritize the Reference. \hl{YEAR}: Describing a calendric year \hl{AGE}: Describing someone's age \hl{CURRENCY}: Reference to some monetary value e.g dollar, euro etc. \hl{PEOPLE}: Describing a single/plural persons \hl{TIME}: Describing a time of the day. Usually you can add the word o'clock after those numbers. \hl{OTHER}: Some other option, which isn't listed here.
 \\ \hline
 Output Space &
 Infinite Set: ["YEAR", "AGE", "CURRENCY", "PEOPLE", "TIME", "OTHER", "REFERENCE phone number", "REFERENCE crooler" ... $\infty$]
 \\ \hline
 Simplified Definition&
 "Keyword Label: YEAR. Keyword Label: AGE. Keyword Label: CURRENCY. Keyword Label: PEOPLE. Keyword Label: TIME. Keyword Label: OTHER. Keyword Label: REFERENCE."
 \\ \hline\hline
 \multicolumn{2}{|c|}{No Mentioned} \\
 \hline
 Description &
 For tasks labeled as \textbf{No Mentioned}, 
 the task definition does not describe the output space by providing keyword labels.
 \\
 \hline
 Original Definition &
 In this task, you're expected to write answers to questions involving multiple references to the same entity. The answer to the question should be unambiguous and a phrase in the paragraph. Most questions can have only one correct answer.
 \\ \hline
 Output Space &
 Infinite Set: [$\infty$]
 \\ \hline
 Simplified Definition &
 ""
 \\ \hline
\end{tabularx}
\caption{
We describe how we created \textit{Simplified} task definition from \textit{Original} task definition for four task definition types: \textbf{Exact Mentioned}, \textbf{Combined Mentioned}, \textbf{Keyword Mentioned}, and \textbf{No Mentioned}. For each task definition type, \textit{Description} describes how the task definition provides the output space information; \textit{Original Definition} shows an example of a task definition within this definition type, which are all retrieved from real tasks in NatInst-V2 dataset; \textit{Output Space} describes the set of the output space; \textit{Simplified Definition} shows an example of how we simplified the Original Task Definition into the simplified version.
}
\label{table:Example-Task-Def}
\end{table*}

\begin{table*}
\centering
\small
\setlength{\tabcolsep}{2mm}
\begin{tabularx}{\linewidth}{|l|X|} 
 \hline
 \multicolumn{2}{|c|}{task036\_qasc\_topic\_word\_to\_generate\_related\_fact} \\
 \hline
 Task Definition & In this task, you need to write a topic word from the given fact. The topic word must have at least one word overlap with the given fact. The topic word often involves adding a new word from a related concept. In your topic word, use at least one word from the given fact. Topic words with two or more words work best.
 \\ \hline
 Task Example & 
 \textbf{Input:} Fact: pesticides cause pollution. \\
 & \textbf{Output:} pollution harms.
 \\ \hline
 \multicolumn{2}{|c|}{task1152\_bard\_analogical\_reasoning\_causation} \\
 \hline
 Task Definition & Two analogies that relate actions with their consequences are given in the form "A : B. C : ?". The phrase "A : B" relates action A to consequence B. Your task is to replace the question mark (?) with the appropriate consquence of the given action C, following the "A : B" relation. Your answer should be a single verb, without further explanation.
 \\ \hline
 Task Example & 
 \textbf{Input:} throw : fly. aspire : ? \\
 & \textbf{Output:} attain
 \\ \hline
 \multicolumn{2}{|c|}{task1159\_bard\_analogical\_reasoning\_containers} \\
 \hline
 Task Definition & Two analogies that relate items to the associated containers is given in the form "A : B. C : ?". "A : B" relates item A to its associated container B. Your task is to replace the question mark (?) with the appropriate container for the given item C, following the "A : B" relation.
 \\ \hline
 Task Example & 
 \textbf{Input:} soda : can. water : ? \\
 & \textbf{Output:} bottle
 \\ \hline
\end{tabularx}
\caption{
We provide several examples of these analogical tasks.
}
\label{table:analogical-example}
\end{table*}

\begin{table}[h!]
\centering
\small
\begin{tabular}{|l|} 
 \hline
task036\_qasc\_topic\_word\_to\_generate\_related\_fact \\
task1152\_bard\_analogical\_reasoning\_causation \\
task1154\_bard\_analogical\_reasoning\_travel \\
task1157\_bard\_analogical\_reasoning\_rooms\_for\_containers \\
task1158\_bard\_analogical\_reasoning\_manipulating\_items \\
task1159\_bard\_analogical\_reasoning\_containers \\
 \hline
\end{tabular}
\caption{List of tasks with the most performance drop when using \textit{Delusive} examples for \textit{Original} model.}
\label{table:analogical-task-list}
\end{table}

\end{document}